\documentclass[12pt]{article}
\usepackage{sbc-template}

\usepackage[utf8]{inputenc}
\usepackage{graphicx,url}
\usepackage{amsmath,amssymb,amsfonts,bm,urwchancal}
\usepackage[mathscr]{euscript}
\usepackage{tikz,standalone}
\usepackage{booktabs}
\usepackage[symbols,record]{glossaries-extra}
\usepackage{xcolor}
\usepackage{hyperref}

\usetikzlibrary{arrows.meta}

\GlsXtrLoadResources[src={symbols},selection=all]

\title{Towards Heterogeneous Multi-Agent Reinforcement Learning with Graph Neural Networks}

\author{Douglas R. Meneghetti\inst{1}, Reinaldo A. C. Bianchi\inst{1}}
\address{Electrical Engineering Department, FEI University Center\\
CEP 09850-901 -- 3972 -- S\~ao Bernardo do Campo -- SP -- Brazil
  \email{\{douglasrizzo,rbianchi\}@fei.edu.br}
}

\begin{document}

\maketitle

\begin{abstract}
	This work proposes a neural network architecture that learns policies for multiple agent classes in a heterogeneous multi-agent reinforcement setting. The proposed network uses directed labeled graph representations for states, encodes feature vectors of different sizes for different entity classes, uses relational graph convolution layers to model different communication channels between entity types and learns distinct policies for different agent classes, sharing parameters wherever possible. Results have shown that specializing the communication channels between entity classes is a promising step to achieve higher performance in environments composed of heterogeneous entities.
\end{abstract}

\section{Introduction}
\label{sec:introduction}

In recent years, multi-agent deep reinforcement learning has emerged as an active area of research. Alongside it, geometric deep learning enables neural networks to perform supervised, semi-supervised and unsupervised learning on data structured as graphs and manifolds. Combining both fields, a new paradigm of multi-agent reinforcement learning has emerged, in which agents learn to communicate \cite{Sukhbaatar2016,Peng2017} by using graph convolution layers as message passing mechanisms \cite{Gilmer2017}.

Until now, new work has focused in the approximation of policies for homogeneous agents, i.e. agents that share the same action set and policy \cite{Agarwal2019,Malysheva2019,Jiang2020}, or in the specialization of agents for a limited number of simple actions \cite{Wang2018b}. However, no work has explicitly studied the potential of creating neural network architectures for environments with heterogeneous agents, capable of specializing the approximated policies according to an agent's class or role in the environment. Such environments may contain heterogeneous teams of agents (\emph{e.g.} drones and terrestrial robots) or homogeneous teams of agents with the need for specialized policies (\emph{e.g.} the RoboCup Soccer Leagues).

In this work, we tackle the challenge of heterogeneous multi-agent reinforcement learning by proposing a neural network architecture that employs information regarding the classes of agents and environment entities to model specialized communication mechanisms, as well as harvest the information regarding agent classes in a heterogeneous multi-agent environment to specialize their communication through the use of inter-class relational graph convolutions.

The text is organized as follows: section~\ref{sec:background} presents the theoretical background in reinforcement learning and graph neural networks; section~\ref{sec:related} presents related work; in section~\ref{sec:proposal}, we introduce the heterogeneous multi-agent graph network, our proposed neural network architecture; sections~\ref{sec:experiments} and \ref{sec:results} present our experiments and results in the StarCraft Multi-Agent Challenge environments and section~\ref{sec:conclusion} concludes the paper.

\section{Research Background}
\label{sec:background}

Reinforcement learning techniques solve tasks that are formalized as Markov Decision Processes (MDPs). An MDP is a tuple \(\langle \gls{S},\gls{actions},\gls{transition},\gls{reward_function} \rangle\), where \(\gls{S}\) is the set of possible states, \gls{actions} the set of actions an agent can perform, \(\gls{transition}: \gls{S} \times \gls{actions} \times \gls{S}\) a state transition function, where \(\gls{transition}(s,a,s')\) maps the probability of an agent observing state \(s'\) after executing action \(a\) in state \(s\). \(\gls{reward_function}: \gls{S} \times \gls{actions} \to \gls{reais} \) is a reward function and \(0 \leq \gamma < 1\) is a discount factor for future rewards, compared to present ones.

Many authors \cite{Littman1994,Bowling2000,Busoniu2008a} propose the modeling of multi-agent systems as stochastic games, which can be considered a generalization of MDPs. In a stochastic game, the set of actions becomes \(\gls{actions} = \gls{actions}_1 \times \gls{actions}_2 \times \ldots \times \gls{actions}_m\) from \(m\) agents; the transition function becomes conditioned on the joint action of all agents, \(\gls{transition}: \gls{S} \times \gls{actions}_1 \times \gls{actions}_2 \times \ldots \times \gls{actions}_m \times \gls{S}\); and the reward function may be different for each agent.

Furthermore, earlier works that have represented MDPs as sets of objects belonging to multiple classes include relational MDPs \cite{Guestrin2003}, object-oriented MDPs (OO-MDPs) \cite{Wasser2008} and multi-agent OO-MDPS \cite{daSilva2019}.

\subsection{Graph Neural Networks}
\label{sec:gnns}

In the same way that successful neural network architectures are biased with relation to the underlying structure of their input data (\emph{e.g.} convolutional neural networks for data with spatial relations and recurrent neural networks for sequential data), the existence of many kinds of data that can be naturally represented as graphs, such as road maps, academic citations \cite{Kipf2017} and molecules \cite{Duvenaud2015}, have prompted the creation of neural network architectures specialized in dealing with graphs.

% TODO talvez essa definição não seja a melhor
A graph \(\gls{graph}\) is composed of a non-empty set of nodes or vertices, denoted as \(\gls{vertices}\), and a set of edges, denoted as \(\gls{edges}\). Each edge \(e \in \gls{edges}\) connects a pair of (not necessarily distinct) nodes \cite{Bondy2008}. When dealing with graphs for the purposes of machine learning, each node, edge and the graph itself may possess features, stored in vectors \cite{Battaglia2018}. \(\vec{v}_i\), \(\vec{e}_j\) and \(\vec{u}\) are the attribute vectors of node \(i\), edge \(j\) and graph \(\gls{graph}\), respectively.

For this work, a graph is defined as a tuple \(\gls{graph} = (\gls{vertices},\gls{edges})\), where vertices in \(\gls{vertices}\) have features vectors and \(\gls{edges}\) is a set of arcs (directed edges) which do not have features.

In its most essential form \cite{Gori2005,Scarselli2009a,Scarselli2009}, a graph neural network allows each node \(i \in \gls{vertices}\) in an input graph to aggregate information from its in-neighbors \(\gls{inneighbors}_{(i)}\), an operation called message passing. Message passing can be expressed generically as \[\vec{u}_i = Agg_{j \in \gls{inneighbors}_{(i)}} \left( f^{\,(l)}\left(\vec{v}_i^{\,(l-1)}, \vec{v}_j^{\,(l-1)},\vec{e}_{(j,i)}\right) \right),\] where \(\vec{v}_i^{\,(l-1)}\) is the feature vector of node \(i\) in layer \(l-1\) of the network, \(\vec{e}_{(j,i)}\) is the feature vector of edge \(e_{(i,j)}\), \(f\) is a parametric transition function that takes into account the state of node \(i\) and its in-neighbors, and \(Agg\) is a permutation-invariant aggregation function, such as average, max or sum.

After the message passing step, the vector of aggregated information \(\vec{u}_i\) is used to generate the output of node \(i\) for layer \(l\) using a parameterized update or output function \(g\), \[\vec{v}_i^{\,(l)} = g^{(l)} \left( \vec{v}_i^{\,(l-1)}, \vec{u}_i \right).\]

Although each node only aggregates information from its in-neighbors, its output can still be influenced by nodes at greater distances by incorporating multiple layers with the aforementioned steps, achieving what can be compared to multi-hop communication.

\section{Related Work}
\label{sec:related}

Works that directly represent multi-agent systems as graphs of agents include DGN \cite{Jiang2020}, MAGNet \cite{Malysheva2019}, NerveNet \cite{Wang2018c} and \cite{Agarwal2019a}. DGN \cite{Jiang2020} introduced the use of graph convolutional layers for inter-agent communication, as well techniques to stabilize training when using these convolutions in RL tasks. The work of \cite{Agarwal2019a} was the first to add non-agent entities to graphs. MAGNet introduces a graph generation layer, which generates the adjacency matrix between agents. NerveNet \cite{Wang2018b} is the first work that tackles the problem of agents of multiple types but, since they work with graphs of fixed sizes and all agents have a single real action, the network does not specialize to these different types of agents.

Preceding work that can be viewed as inter-agent communication with GNNs include CommNet \cite{Sukhbaatar2016} and BiCNet \cite{Peng2017}. More recent work that also focuses on modulating when agents communicate with each other include ATOC \cite{Jiang2018} and TarMAC \cite{Das2019}.

This work differs from previous ones by making use of node class information both as a means to specialize communication between node classes as well as for learning a centralized policy for multiple agents of the same class.

\section{Heterogeneous Multi-agent Graph \gls{Q}-Network}
\label{sec:proposal}

In this work, states are represented as directed labeled graphs, in which nodes represent either agents or environment entities; arcs represent communication channels either among agents or between an agent and an environment entity; node labels represent agents/entities classes and edge labels represent specialized communication channels. In practical settings, the existence of an arc between a node \(v\) and an agent \(z\) may be related to \(z\)'s capability of observing \(v\) and an arc from agent \(z_1\) to agent \(z_2\) indicate an open communication line from \(z_1\) to \(z_2\).

For a graph \(\gls{graph}\), each node \(v \in \gls{vertices}(\gls{graph})\) is associated with a node class \(c \in \gls{C}\), where \(\gls{C}\) is the set of node classes. The class of a node \(v\) can be accessed through a function \(\gls{C}(v)\). A subset \(\gls{Z}\) of \(\gls{C}\) contains agent classes, \emph{i.e.} classes pertaining solely to agent nodes.

The class of a node determines the number of state variables used to describe that node. Furthermore, the class of an agent \(z\) determines its action set \(\gls{actions}_{\gls{C}(z)}\), as well as its policy \(\gls{pi}_{\gls{C}(z)}\). In our work, arcs are used to encode relations between node classes. More specifically, an arc labeled \((n, m)\) represents a relation between an agent of class \(n\) with another node of class \(m\).

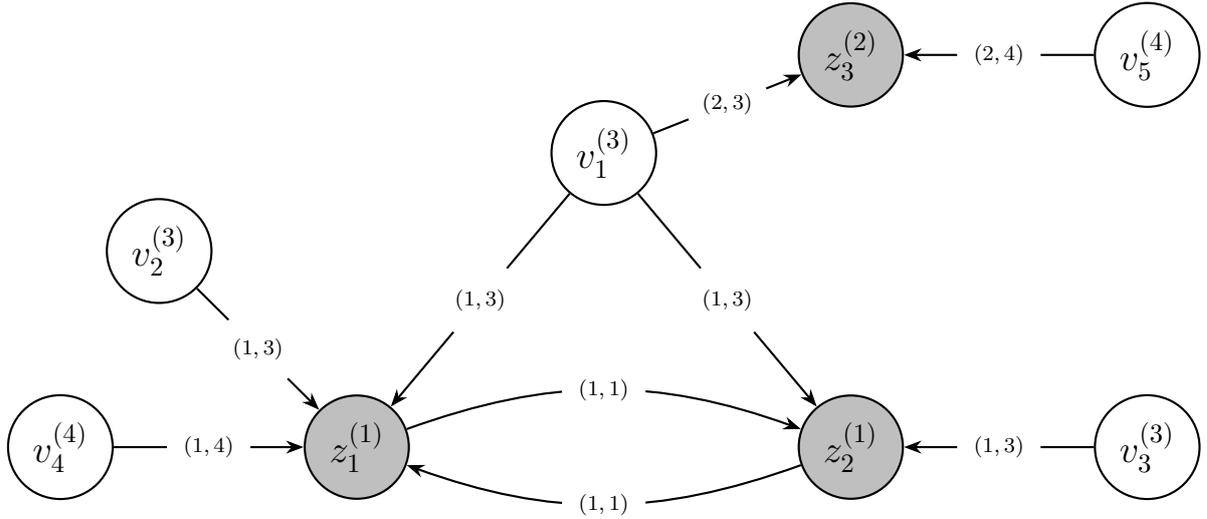
\begin{figure}
	\centering
	\begin{tikzpicture}
    \begin{scope}[every node/.style={fill=lightgray,circle,thick,draw,font=\large}]
        \node (z1) at (0, 0) {\(z_1^{(1)}\)};
        \node (z2) at (6.5, 0) {\(z_2^{(1)}\)};
        \node (z3) at (6.5, 5.2) {\(z_3^{(2)}\)};
    \end{scope}

    \begin{scope}[every node/.style={circle,thick,draw,font=\large}]
        \node (v1) at (3.25, 3.9) {\(v_1^{(3)}\)};
        \node (v2) at (-2.6, 2.6) {\(v_2^{(3)}\)};
        \node (v3) at (10.4, 0)  {\(v_3^{(3)}\)};
        \node (v4) at (-3.9, 0) {\(v_4^{(4)}\)};
        \node (v5) at (10.4,5.2)   {\(v_5^{(4)}\)};
    \end{scope}

    \begin{scope}[>={Stealth[black]},
        every node/.style={fill=white,circle,font=\scriptsize},
        every edge/.style={draw=black,thick}]
        \path [->] (z1) edge  [bend left=20] node {\((1,1)\)} (z2);
        \path [->] (z2) edge  [bend left=20] node {\((1,1)\)} (z1);
        \path [->] (v1) edge node {\((1,3)\)} (z1);
        \path [->] (v2) edge node {\((1,3)\)} (z1);
        \path [->] (v4) edge node {\((1,4)\)} (z1);
        \path [->] (v1) edge node {\((1,3)\)} (z2);
        \path [->] (v3) edge node {\((1,3)\)} (z2);
        \path [->] (v1) edge node {\((2,3)\)} (z3);
        \path [->] (v5) edge node {\((2,4)\)} (z3);
    \end{scope}
\end{tikzpicture}
	\caption{A multi-agent system represented as a graph.}
	\label{fig:graph-example}
\end{figure}

Figure \ref{fig:graph-example} exemplifies a graph with three agents and five environment entities, in which agent nodes aggregate information from their neighbors. In the figure, \(v_i^{(j)}\) represents node \(v\) with index \(i\) and class \(j\). Darker nodes represent agents, while lighter nodes represent other environment objects encoded in the graph state.

\subsection{Neural Network Architecture}

The proposed neural network architecture, denominated Heterogeneous Multi-Agent Graph \gls{Q}-Network (HMAGQ-Net), is composed of three modules: an encoding module, a communication module and an action selection module. The modules are applied in sequence to the input data and the model is capable of being trained end-to-end. The full network architecture is presented in figure~\ref{fig:hmagqnet} and explained below.

\subsubsection{Encoding}

In order to deal with the varying number and meaning of state variables that compose each node class, we introduce an encoding function \(\phi_c\) for each \(c \in \gls{C}\), which receives as input the vector \(\vec{v} \in \gls{reais}^{d_{\gls{C}(v)}}\) containing a node's description, and outputs an encoded vector \(\phi_c(\vec{v}) \in \gls{reais}^m\), where \(m\) is a common output size for the encoding functions of all classes. In this work, we explore using multi-layer perceptrons as implementations of \(\phi\).

\subsubsection{Communication}

In the communication layer, each agent node \(z\) aggregates information from its set of in-neighbors nodes \(\gls{inneighbors}_{(z)}\). In HMAGQ-Net, we employ relational graph convolutions (RGCN) \cite{Schlichtkrull2018} to allow for specialization of the message passing mechanism. In an RGCN layer, the feature vector of node \(i\) in layer \(l+1\) is given by \[ \vec{v}_i^{\,(l+1)} = \sigma \left( \sum_{r \in \gls{relations}} \sum_{j \in \mathscr{N}^{r}_{i}} \frac{1}{c_{i,\,r}} \bm{W}^{(l)}_r \vec{v}^{\,(l)}_j + \bm{W}^{(l)}_0 \vec{v}^{\,(l)}_i \right), \] where \(r\) represents the index of a relation between nodes \(i\) and \(j\). In this work, the set of relations is defined as all possible pairs \((c_1, c_2)\), \(c_1 \in Z\), \(c_2 \in C\) (see arc labels in figure \ref{fig:graph-example}).

Regularization in RGCN is achieved by decomposing parameter matrix \(\bm{W}^{(l)}_r\) into \(B\) basis transformations \(\bm{V}\) and coefficient vectors \(a\), \[ \bm{W}^{(l)}_r = \sum_{b=1}^B \vec{a}^{\,(l)}_{rb} \bm{V}^{(l)}_b.\] In this way, all relations \(r \in \gls{relations}\) share the same set of basis matrices, while coefficient vectors depend on \(r\). In our work, the number of relations is \(|\gls{relations}| = |\gls{Z}| \cdot |\gls{C}|\), as each agent class models a specialized communication channel with all  other node classes.

% \paragraph{Attentional communication}

% Recent successful work in multi-agent reinforcement learning using graph neural networks \cite{Jiang2020,Agarwal2019a,Wang2018b} employ some kind of self-attention mechanism in their architecture. To that end, we employ graph attention (GAT) layers \cite{Velickovic2018} in the communication module. In a GAT layer, the output vector of node \(i\) at layer \(l+1\) is given by equation~\eqref{eq:gat-output}, where \(\alpha_{ij}\) is a self-attention coefficient, given as the softmax output of a single layer perceptron, which receives as input the concatenation of feature vectors from nodes \(i\) and \(j\) and normalizes values across all in-neighbors of \(i\).

% In equations \eqref{eq:gat-output} and \eqref{eq:gat-attention}, \(M\) refers to the number of self-attention heads, each one produced by a different single layer perceptron, potentially augmenting the expressiveness of the attention mechanism; and \(\|\) refers to the vector concatenation operation.

% \begin{gather}
%     \label{eq:gat-output}
%     \vec{v}_{i}^{\,(l+1)} = \|_{m}^{M} \sigma \left( \sum_{j \in \gls{inneighbors}_i} \alpha_{ij}^{m} W^m \vec{v}_{j}^{\,(l)} \right) \\
%     \label{eq:gat-attention}
%     \alpha_{ij} =
%     \frac{\exp \left( \mathrm{LeakyReLU} \left( \vec{a}^{\,\top}[\bm{W}\vec{v}_i \, \| \, \bm{W}\vec{v}_j] \right) \right)}
%     {\sum_{j \in \mathscr{N}(i) \cup \{ i \}} \exp\left( \mathrm{LeakyReLU} \left( \vec{a}^{\,\top} [ \bm{W}\vec{v}_i \, \| \, \bm{W}\vec{v}_j ] \right) \right)}
% \end{gather}

\subsubsection{Action selection}

After \(K\) layers of graph convolutions, the final feature vectors of the agent nodes are taken as their individual observations of the graph. We introduce a function \(\gls{Qc}\) for each agent class \(c \in \gls{Z}\), which receives the observation \(o_z\) of an agent \(z\) of class \(c\) as input and outputs a vector of size \(|A_c|\), corresponding to the observation-action values for agent \(z\).

\begin{figure}
	\centering
	\includegraphics[width=\textwidth]{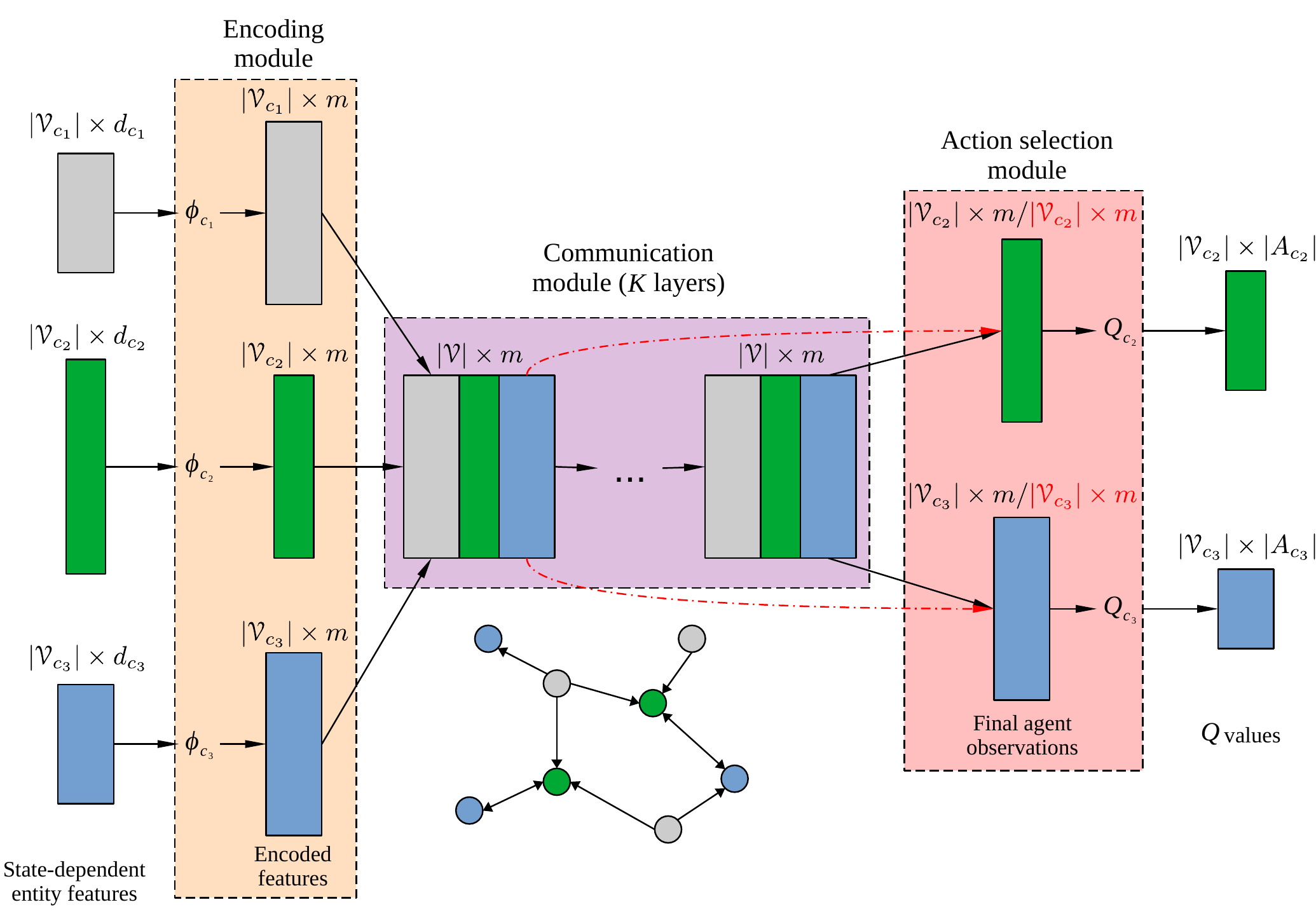}
	\caption{An example of the proposed model processing a graph of 3 environment entities of class \(c_1\) (gray), 2 agents of class \(c_2\) (green) and 3 agents of class \(c_3\) (blue). Red elements refer to changes in the network topology if the output of all \(K\) layers from the communication module are used as input for the action module.}
	\label{fig:hmagqnet}
\end{figure}

Optionally, the concatenation of the feature vectors generated by all graph convolution layers may be taken as the final observation for each agent \cite{Jiang2020}, an alternative named in the experiments as ``full receptive field'' and displayed as red arrows in figure~\ref{fig:hmagqnet}.

\subsection{Training stabilization}

We employ both a policy network and a target network, with the same topology. The target network is responsible for generating stable targets and is updated with a copy of the parameters of the policy network after a fixed number of time steps. The parameters of the policy network are optimized during every step of the environment with a batch of transitions sampled from a replay buffer.

To speed up training, proportional prioritized experience replay \cite{Schaul2015a} was implemented, in which each transition of the replay buffer maintains a tuple \(\langle s,\vec{a},s',\vec{r} \rangle\), where \(s\) and \(s'\) are states represented as graphs, \(\vec{a}\) are the actions selected by all agents and \(\vec{r}\) are the rewards observed by each agent.

The loss function is given by \[ J(\theta) = \sum_{c \in Z} \frac{1}{|Z|} \sum_{z \in c} (r_i + \gamma \max_{a'} \gls{Qc}(o'_{z}, a'; \theta^-) - \gls{Qc}(o_z, a_z; \theta))^2, \] where \(Z\) represents all agent classes, \(z\) is a single agent, \(\theta\) are the parameters of the policy network and \(\theta^-\) are the parameters of the target network.

% Further training stabilization is achieved by clipping the global gradient norm of the networks weights.

% A final component of the loss function is temporal relation regularization \cite{Jiang2020}, a term which attempts to approximate the attention weights of the last layer of the communication module when applied to two subsequent states. This is done by including the Kullback-Leibler divergence between the two distributions of the attention weights in the final loss function of both methods. Let \(\mathcal{G}(s;\theta)\) denote this distribution when the network observes state \(s\). The final loss function used by the neural network is given by \[ \mathcal{L}(\theta) = J(\theta) + \lambda_{\text{trr}} D_{KL}(\mathcal{G}(s;\theta)\|\mathcal{G}(s';\theta)), \] where \(\lambda_{\text{trr}}\) is a coefficient used to tune the contribution of temporal relation regularization to the final loss.

\section{Experiments}
\label{sec:experiments}

The proposed model was tested in the StarCraft Multi-Agent Challenge (SMAC) domain \cite{Samvelyan2019}, a collection of maps for the StarCraft II Learning Environment focused in multi-agent tasks. In the maps, \(n\) units from the player team are individually controlled in order to achieve victory in a battle scenario against \(m\) units from the adversary team. Each unit belongs to one of multiple classes, which may be described by different state variables and have different action sets and optimal policies. In each of the maps, each node class possesses between 4 and 6 features. Agent classes have 4 movement actions, \(m\) attack actions and 1 no-op action for incapacitated units (all discrete). Since units have different behavior (movement speed, attack range) it is expected that learning different policies for each unit type will be beneficial for the player team.

\begin{table}
	\centering
	\caption{Hyperparameters used in the training setting}
	\label{tbl:hyperparams}
	\begin{tabular}{rl}
		Training steps                 & \(10^6\)        \\
		\(\hat\theta\) update interval & \(250\)         \\
		Network learning rate          & \(2.5*10^{-4}\) \\
		L2 regularization coef.        & \(10^{-5}\)     \\
		TRR coef.                      & \(0.01\)        \\
		RL discount factor \(\gamma\)  & \(0.99\)        \\
		\(\epsilon_{max}\)             & \(0.95\)        \\
		\(\epsilon_{min}\)             & \(0.1\)         \\
		Proportional PER \(\alpha\)    & \(0.6\)         \\
		Proportional PER \(\beta\)     & \(0.4\)
	\end{tabular}
\end{table}

In all tests, each network \(\phi\) in the encoding layer was an MLP with two hidden layers of 128 neurons and an output encoding of 64 values. The communication module was composed of 4 relational layers, with the first layer having an input vector of 64 values, the last layer having an output vector of 64 values, and all hidden connections being composed of vectors of 128 values. The relational module was tested against an attentional communication module composed of graph attention layers \cite{Velickovic2018}. The attention layers worked with 4 attention heads, whose output was concatenated at the end of each layer. Finally, the \gls{Q} networks for agent classes were MLPs with 64 values in the input layer, two hidden layers with 128 neurons and output vector size equal to the number of actions of each agent class. For all modules, the sigmoid nonlinearity was used, as well as the Adam optimizer. Hyperparameters are provided in table~\ref{tbl:hyperparams}.

Additional experiments were performed to evaluate the performance of using the full receptive field (FRF) as the final agent observations; giving the agents the ability to communicate by creating arcs between them, regardless of distance (full agent communication, FAC) and the use of temporal relation regularization in the attentional model our proposal was tested against (TRR, \cite{Jiang2020}).

Experiments were performed in a mix hardware environment, a computer equipped with an Nvidia GTX 1070 and a server equipped with an Nvidia V100. Each run took an average time of 70 hours to complete.

\section{Results}
\label{sec:results}

Table~\ref{tbl:results} displays the results of the different trained models. Two values were taken as measures of performance for the agent team: final episode reward and number of steps the agent team remained alive. In the SMAC environments, agents with larger rewards were able to deal more damage to the opponent teams, while longer episodes indicate agents that were able to survive for longer.

When tested against a random baseline (a group of agents which only take random actions), all the trained models had superior performance in both measures. The two models that accumulated the most average reward by episode employed RGCN layers, while the model that remained alive for the most number of steps was a GAT model.

\begin{table}
	\centering
	\caption{Results of applying HMAGQ-Net on the 2s3z map of the SMAC domain under different configurations. FRF = full receptive field. FAC = full agent communication. TRR = temporal relation regularization.}
	\label{tbl:results}
	\begin{tabular}{cccc|cc|cc}
		\multicolumn{4}{c|}{}                & \multicolumn{2}{c|}{Mean n. steps} & \multicolumn{2}{c}{Mean reward}                                                                                      \\
		Comms module                         & FRF                                & FAC                             & TRR              & All            & Last 10\%      & All           & Last 10\%     \\ \midrule
		RGCN                                 & \( \checkmark \)                   & \( \checkmark \)                &                  & \(77.66\)      & \(82.18\)      & \(\bm{4.69}\) & \(\bm{4.79}\) \\
		RGCN                                 & \( \checkmark \)                   &                                 &                  & \(78.65\)      & \(83.62\)      & \(3.82\)      & \(3.77\)      \\
		RGCN                                 &                                    &                                 &                  & \(76.85\)      & \(81.93\)      & \(4.24\)      & \(4.30\)      \\
		GAT                                  & \( \checkmark \)                   & \( \checkmark \)                & \( \checkmark \) & \(70.63\)      & \(75.10\)      & \(3.85\)      & \(3.86\)      \\
		GAT                                  & \( \checkmark \)                   &                                 & \( \checkmark \) & \(77.20\)      & \(82.13\)      & \(3.53\)      & \(3.47\)      \\
		GAT                                  &                                    & \( \checkmark \)                & \( \checkmark \) & \(\bm{79.41}\) & \(\bm{84.59}\) & \(3.97\)      & \(3.99\)      \\
		GAT                                  &                                    &                                 &                  & \(77.21\)      & \(82.29\)      & \(3.98\)      & \(3.99\)      \\ \midrule
		\multicolumn{4}{c|}{Random baseline} & \multicolumn{2}{c|}{\(52.155\)}    & \multicolumn{2}{c}{\(2.222\)}
	\end{tabular}
\end{table}

Overall, networks that employed full-agent communication (FAC) achieved a higher reward than networks that did not. However, it is hard to observe better performance of agents that received the full receptive field (FRF) of the communication layers as their observations or used TRR in their loss function. This may be due to the fact that these techniques were first proposed to solve the simpler predator-prey environment, with homogeneous agents \cite{Jiang2020}.

\begin{figure}[!htb]
	\centering
	\includegraphics[width=\textwidth]{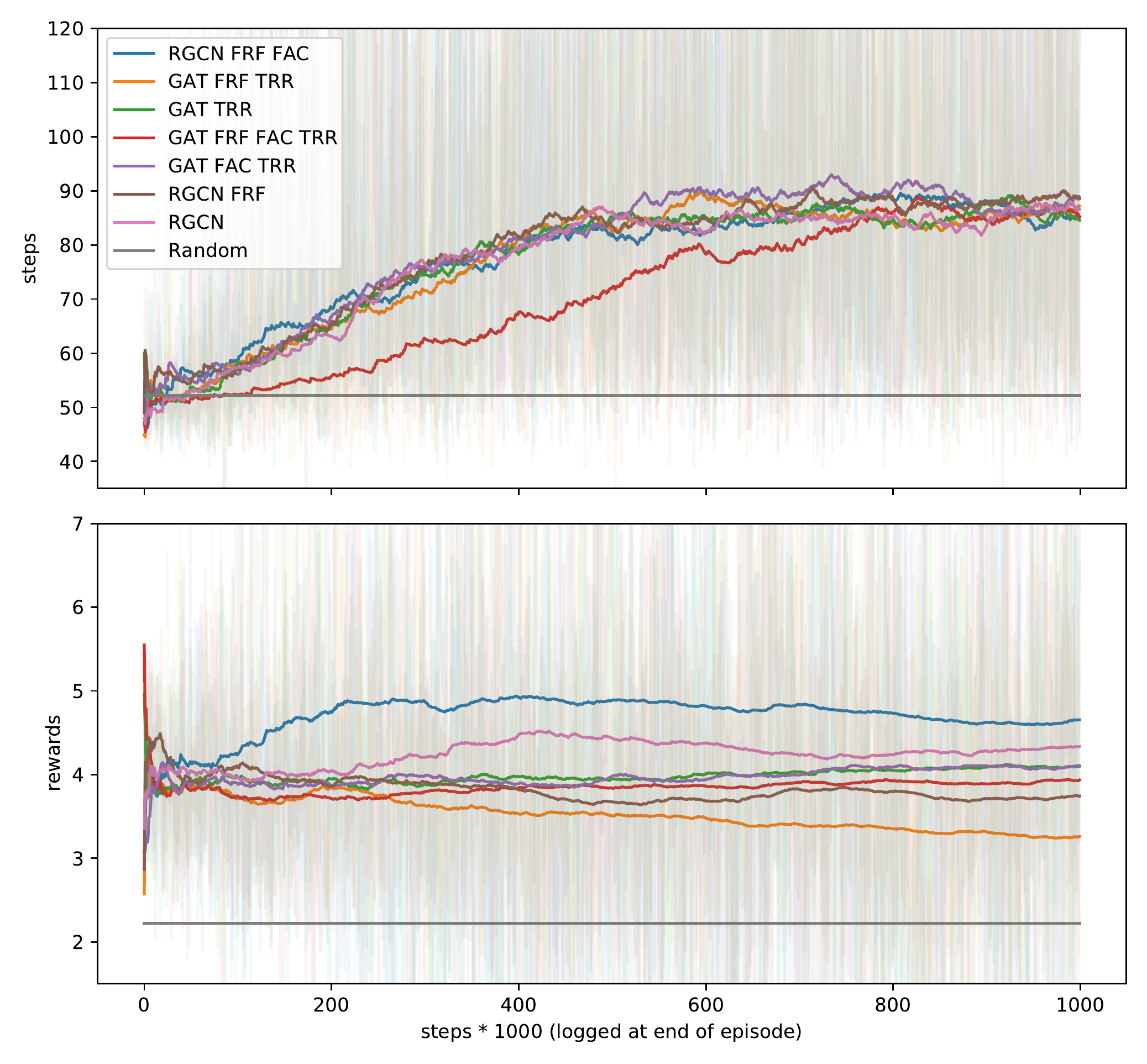}
	\caption{Number of steps (top) and average reward collected by each agent (bottom), per episode, for a total of 1 million training steps.}
	\label{fig:results}
\end{figure}

Figure~\ref{fig:results} presents the same measures throughout the training. Since the measures were noisy, we employed exponential smoothing to better visualize the trend lines. It can be seen from the top graph that all models tended to converge to the same number of steps alive, while achieving different final rewards. It can also be seen that the two RGCN models that accumulated the most reward dominated all other models in that measure through most of the training time.

\section{Conclusion}
\label{sec:conclusion}

This work presented the Heterogeneous Multi-agent Graph \gls{Q}-Network, a neural network architecture that processes environment states represented as directed labeled graphs and employs relational graph convolution layers to achieve specialized communication between agents of heterogeneous classes, as well as multiple encoding networks to normalize entity representation and multiple action networks to learn individual policies for each agent class.

Results have shown that specializing the communication channels between entity classes is a promising step to achieve higher performance in environments composed of heterogeneous entities. In future work, we intend to test HMAGQ-Net on multiple environments with different number of agents and agent classes; isolate the contribution of learning policies for agent classes by testing variants which learn a single policy for all agents and individual policies for each agent; and propose an action module trained via policy gradient.

\section*{Acknowledgments}

The authors acknowledge the S\~ao Paulo Research Foundation (FAPESP Grant 2019/07665-4) for supporting this project. This study was financed in part by the Coordenação de Aperfeiçoamento de Pessoal de Nível Superior - Brasil (CAPES) - Finance Code 001.

\end{document}